\documentclass[letterpaper, 10 pt, conference]{ieeeconf}  

\IEEEoverridecommandlockouts                              

\usepackage{graphics} 
\usepackage{balance}
\usepackage[dvipsnames]{xcolor}
\usepackage{lipsum}
\usepackage{tikz, pgfmath}
\usepackage{amsmath,amsfonts,amssymb}
\usepackage{mathtools}
\usepackage{cancel}
\usepackage{url}
\usepackage{booktabs}
\usepackage{subcaption}
\usepackage{multirow}
\usepackage{graphicx}
\usepackage{balance}
\usepackage{multirow}
\usepackage{float}

\allowdisplaybreaks

\newcommand{\Vector}[1]{%
    \ifcat\noexpand#1\relax
        \boldsymbol{#1}
    \else
        \mathbf{#1}
    \fi
}

\newcommand{\Real}{\mathbb R}
\newcommand{\Transpose}[1]{{#1}^{\top}}

\def\figurename{Fig.}

\title{\LARGE \bf
Decentralized UAV Swarms for Ground Target Protection in GPS- and Communication-Denied Environments
}
\author{Dimitria Silveria$^{1}$, Paulo Ricardo Marques de Araujo$^{2}$, Tiago Nascimento$^{3}$ and Sidney Givigi$^{2}$
\thanks{$^{1}$D. Silveria is with the Department of Electrical and Computer Engineering, and the Ingenuity Labs Research Institute, Queen's University, Kingston, ON, Canada,
{\tt\small dimitria.s@queensu.ca}.}%
\thanks{$^{2}$Paulo Ricardo Marques de Araujo and Sidney Givigi are with the School of Computing and the Ingenuity Labs Research Institute, Queen's University, Kingston, ON, Canada, {\tt\small \{paulo.araujo, sidney.givigi\}@queensu.ca}.}%
\thanks{$^{2}$Tiago Nascimento is with the Lab of Systems Engineering and Robotics (LASER), Department of Computer Systems, Universidade Federal da Paraíba, and with the Department of Cybernetics, Czech Technical University in Prague, Prague 16629, Czechia {\tt\small tiagopn@ci.ufpb.br}.}%
}

\begin{document}

\maketitle

\thispagestyle{empty}
\pagestyle{empty}

\begin{abstract}
The presence of UAVs in military operations has recently increased, also increasing the demand for defense systems against UAV attacks. UAVs can also be used as countermeasures. Most available methods rely on UAV-to-UAV communication and global positioning. However, such resources may not be available in modern warfare scenarios.
To address these limitations, we propose a pipeline for ground-target protection against UAV attacks that employs autonomous swarms of UAVs.
We assume a communication- and GPS-denied environment in which the UAVs use onboard sensors to track the target and coordinate as a swarm. We developed Kalman filters to estimate the states of unknown targets and the positions of UAVs in the swarm using only relative measurements.
Also, our strategy is to encircle the target of interest to maximize coverage. To achieve that, we propose a decentralized swarm encirclement technique that adapts to the target's motion.
Our approach was extensively validated using real robots, demonstrating its effectiveness in detecting, encircling, and intercepting hostile UAVs. Code available at \url{https://github.com/QUARRG/target_protection}
\end{abstract}

\section{INTRODUCTION} \label{sec:introduction}

The use of commercial uncrewed autonomous vehicles (UAVs), also known as drones, in military operations has grown significantly over the past decade~\cite{combat-use-commercial-drones_2023}. Compared to crewed aircraft, UAVs are cheaper, lighter, and more agile, and can be used for surveillance, jamming, decoy operations, and strikes~\cite{minculete_approaches-to-combate-the-use-of-drones_2024}. 
In the Ukraine conflict, for example, UAVs have been extensively used to detect and strike ground tanks and troops~\cite{minculete_approaches-to-combate-the-use-of-drones_2024}, highlighting the importance of technologies to protect terrestrial operations against aerial attacks. 

Defending against drone attacks requires reliable detection and interception. 
Common countermeasures include jammers, GPS spoofers, and snaggers~\cite{Cisar_anti-drone-defense_2020}, but these approaches are power-intensive, prone to causing undesired interference with devices other than the intruder drone, and are limited in mobility~\cite{Tesfay_smart-jamming_2024, Rudys_hostile-uav-detection-neutralization_2022}.
Autonomous drone-to-drone interception has recently emerged as a good alternative, especially with swarms of UAVs~\cite{Athanasios_gps-denied-drone-interception_2024, Castrillo_review-counter-uas-technologies_2022, Souli_MAS-drone-interception_2023}.

Protecting ground vehicles against drone swarms is a difficult task, as it requires simultaneous tracking of the protected ground vehicle and detection, interception, and neutralization of hostile UAVs. Most existing works address ground target tracking~\cite{Liu_distance-based-ground-target-encirclement_2025} and aerial interception~\cite{Yang-Cooperative_interception_with_angle_constraints_uavs-2023} separately.
\begin{figure}
    \centering
    \includegraphics[trim={0cm 5cm 7cm 0},clip,width=\columnwidth]{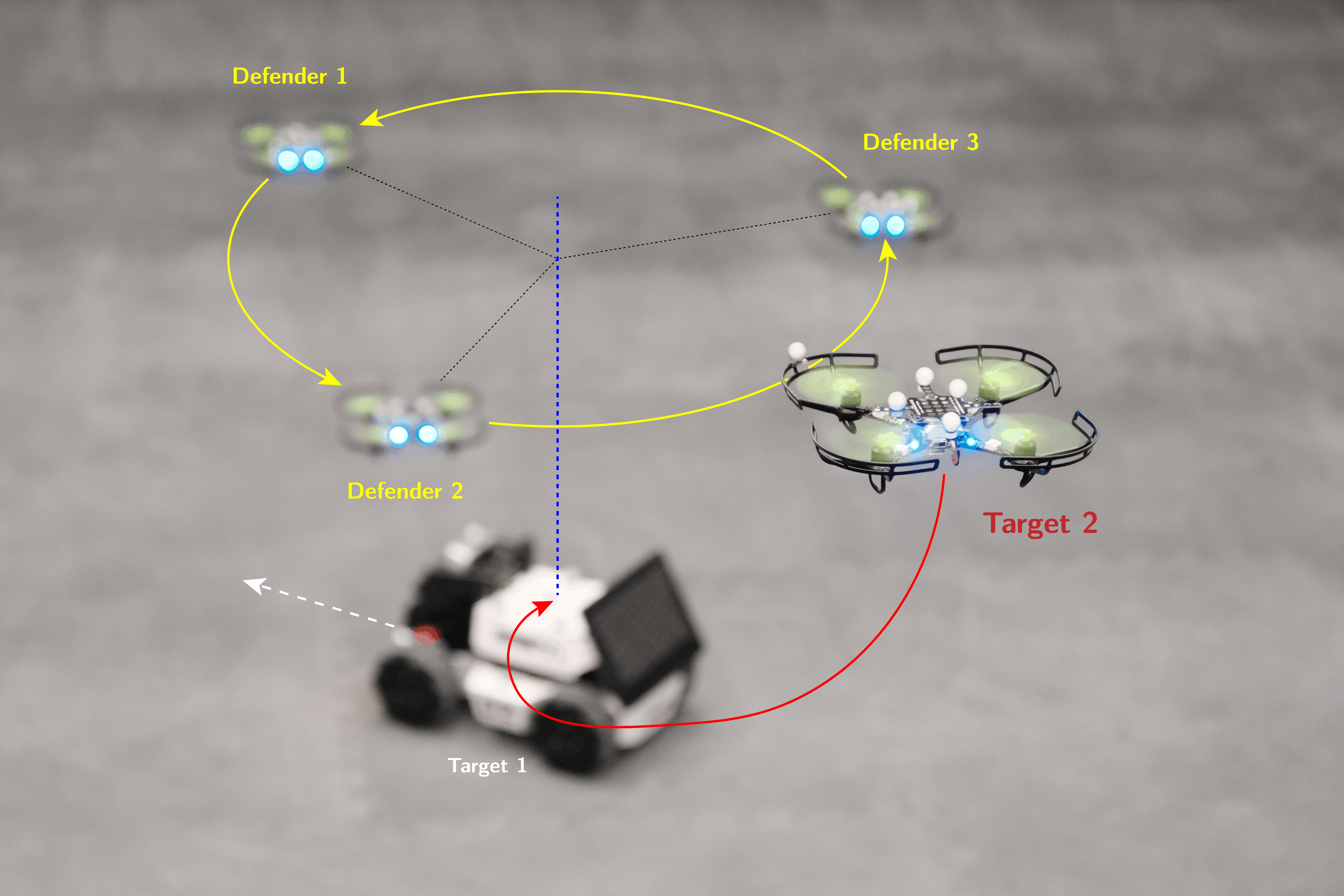}
    \caption{Target defense experimental setup.}
    \label{fig:scenario}
\end{figure}
\begin{figure*}[t]
    \centering
    \includegraphics[width=\textwidth]{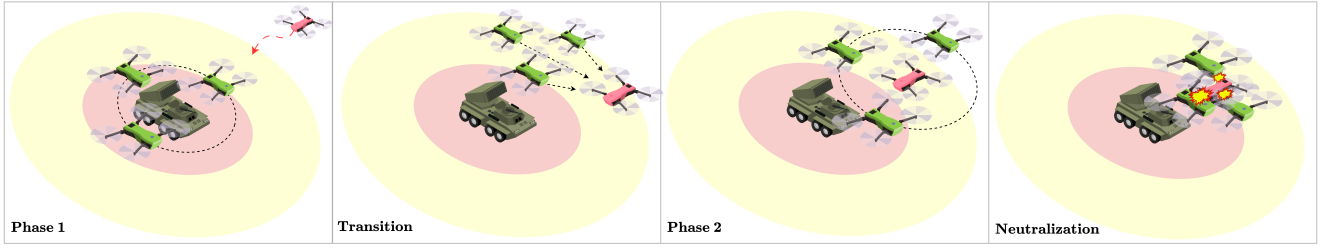}
    \caption{Four phases of the pipeline presented in this paper for ground vehicle encirclement and protection against attacker drones. The green UAVs represent the Defenders, and the red one is the Attacker.}
    \label{fig:pipeline}
\end{figure*}

Many target interception and encirclement approaches assume full, noise-free access to the target's states~\cite{Yang-Cooperative_interception_with_angle_constraints_uavs-2023}, are validated only in simulation~\cite{Zhang-multi_target_encirclement_transformer_rl-2025}, or are centralized and limited to a certain number of agents~\cite{Liu-baseline_iros-2025}. Even more realistic approaches often assume GPS availability, inter-agent communication, or noiseless measurements~\cite{Yang-time_varying_formation_encircling_tracking_control-2025}. 

For example, the target encirclement and tracking control in \cite{Yang-time_varying_formation_encircling_tracking_control-2025} proposes a vision-based framework that estimates the target's position. However, it assumes UAVs have access to GPS sensors and inter-agent communication, and that the measurements are noiseless. The work in~\cite{Ranjan_self-organizing-multiagent_2025}, on the other hand, addresses GPS-denial, requires minimal inter-agent communication, and is decentralized. However, it assumes a stationary target, and it does not fully eliminate inter-agent communication.

Similarly, most cooperative terrestrial target tracking pipelines rely on GPS and/or communication, or are designed for a specific swarm size~\cite{Henati_convoy-ground-protection-uavs_2021, Zhou_surveillance-coverage-ground-vehicle-collaborative_2024,Sivakumar_mpc-path-planning-convoy-protection_2021, Liu_distance-based-ground-target-encirclement_2025}. 

Those assumptions are not realistic in real warfare environments, where communication may be jammed, GPS spoofed, and sensor measurements corrupted by noise. Although some studies address communication-denied~\cite{Zhou_state-estimation-swarm-communication-deprivd_2025, Zhou_decentralized-quad-swarm-communication-denied_2021} or GPS-denied scenarios~\cite{Zhang_network-navigation-algorithm_2021,Goudar_decentralized-onboard-range-aided-coop-localization-communication-denied_2026}, fully integrated protection pipelines under such constraints remain widely unexplored. 


To our knowledge, the only recent work addressing the entire pipeline of following a land vehicle, detecting, tracking, and neutralizing a hostile drone using drones is~\cite{Liu-baseline_iros-2025}. While it estimates the attacker's position without GPS using noisy sensor measurements, it is limited to only two guardian drones, assumes inter-agent communication, and requires the ground vehicle to transmit its position to the guardian drones at all times,
which is usually not the case in warfare scenarios.
These constraints reduce its applicability in realistic combat scenarios.

To address some of these issues, this paper provides a novel decentralized, swarm-based pipeline for autonomous protection of ground vehicles. A UAV swarm encircles a moving ground vehicle while monitoring for unauthorized drones. Upon detection, the swarm encircles the intruder to assess the threat and, if necessary, neutralizes it by collapsing the encirclement region. This strategy increases robustness, as neutralization does not depend on a single UAV.

To perform the target encirclement, we employ the technique proposed in~\cite{swarm_embedding}, which is decentralized and scalable. We have improved this pipeline by adapting it to moving targets and to operate solely with noisy relative position measurements, thereby eliminating the need for global positioning and inter-agent communication.  
To the best of our knowledge, this is the first work to demonstrate a complete, communication-free, and GPS-denied pipeline for autonomous drone-ground vehicle protection and drone-to-drone interception.

Our key contributions are:
\begin{enumerate}
    \item An Invariant Extended Kalman filter to estimate the states and control inputs of a ground or aerial target in an arbitrary coordinate system using only relative measurements that are obtained by onboard sensors;
    \item A discrete Kalman filter to estimate the phase difference between agents flying in a circular formation using only relative onboard measurements;
    \item An adaptive swarm coordination algorithm for target encirclement. The algorithm adapts to the velocity of the target to adjust the area covered by the swarm to keep the target under containment;
    \item Extensive real-robot experiments, with three UAVs and one ground vehicle to validate the proposed approach.
\end{enumerate}


\section{PROBLEM FORMULATION} \label{sec:problem-formulation}
We consider the problem depicted in \figurename~\ref{fig:scenario} of protecting a ground target, \textbf{Target 1}, using a swarm of $n$ UAVs (\textbf{Defenders}) arranged in an encirclement formation to maximize the coverage area. \textbf{Target 1} is attacked by a hostile UAV, \textbf{Target 2}.

To emulate warfare conditions, the \textbf{Defenders} have no prior knowledge of the environment, no GPS, and no inter-agent communication. They rely solely on onboard odometry (e.g., radar- or camera-based if active sensors pose risks to safe operation) for self-localization. 
We assume that low-level pose estimation and target detection modules are available; their design is outside the scope of this work. Neither the \textbf{Defenders} nor \textbf{Target 1} exchange information.



The problem is divided into four different phases, as illustrated in \figurename~\ref{fig:pipeline}. In Phase 1, the \textbf{Defenders} encircle \textbf{Target 1} on a fixed horizontal plane while estimating its motion using only noisy relative measurements. 
Similarly, swarm coordination has to be performed through relative measurements from the \textbf{Defenders'} onboard sensors to keep \textbf{Target 1} at the center of the encirclement.

When \textbf{Target 2} enters the \textbf{Defenders'} detection radius $r_{\text{detec}}$ (the \textit{Yellow Zone}). a \textbf{Transition Phase} begins. The \textbf{Defenders} use a flocking algorithm based on noiseless relative positions to \textbf{Target 2} and neighboring UAVs to approach the attacker (\textbf{Target 2}) while avoiding collisions. 
After that, Phase 2 begins, and the \textbf{Defenders} encircle \textbf{Target 2} similarly to \textbf{Target 1}, but the encirclement plane varies according to \textbf{Target 2}'s altitude. A \textit{Red Zone} of radius $r_\text{safe}$ is defined around \textbf{Target 1}. If \textbf{Target 2} is outside this zone, it is not perceived as a threat. Upon entering this zone, it becomes a threat, and the encirclement triggers the Neutralization Phase, in which they collapse onto the attacker. If \textbf{Target 2} leaves the \textit{Yellow Zone}, it is no longer a target, and the \textbf{Defenders} return to Phase 1. 

\section{PROPOSED METHODS}

\subsection{Swarm Encirclement Control}\label{sub-sec:swarm-control}
We propose the following method to coordinate a swarm of UAVs to encircle a point moving in space.

Consider a swarm of $n$ UAVs moving in a circular trajectory of radius $r(t)$ and angular velocity $\omega_{z,d}(t)$ around a target. The encirclement swarm control proposed in this section acts on each UAV's angular velocity to ensure that all UAVs' angular separation is uniform, that is, $360^\circ/n$, avoiding inter-drone collision.

Given that $\phi^i(t)$, $\phi^k(t)$, and $\phi^j(t)$ are the phases of ego, leader, and follower UAVs, we can apply the phase separation control law proposed in~\cite{swarm_embedding}.
This controller uses these phases to calculate the desired angular separation that maintains the uniform phase separation:
\begin{equation}
        \omega^i_{z}(t+1) = \omega_{z,d}(t) + k_\phi \left(\frac{1}{\phi^{ki}(t)} + \frac{1}{\phi^{ji}(t)}\right),
    \label{eq:phase-control}
\end{equation}
where $\omega^i_{z}(t+1)$ is the desired angular velocity of the ego UAV, in the next time step, $\phi^{ki}(t)=\phi^i(t)-\phi^k(t)$, $\phi^{ji}(t)=\phi^i(t)-\phi^j(t)$, $\phi^{ki}, \phi^{ji} \in [-\pi,\pi]$, and $k_\phi > 0$ is the gain. When all agents achieve uniform separation, they all move at the desired nominal angular velocity $\omega_{z,d}(t)$.

Next, $\omega^i_{z}(t+1)$ is converted into an incremental rotation matrix using the exponential operator, according to
\begin{equation}
    \label{eq:theta-dynamics}
    \Delta R_{c}^i(t+1) = \exp(\omega^i_{z}(t+1)\Delta t [\Vector{e}_z]_\times),
\end{equation}
with $\Vector{e}_z=[0 \text{ }0\text{ }1]^\top$.

The next position of the UAV is computed as:
\begin{equation}\label{eq:radial-position}
    \Vector{x}^i(\phi^i(t+1)) = \Delta R_{c}^i(t+1) \Vector{p}_c,
\end{equation}
where $\Vector{p}_c = r(t) \Transpose{\left[ \cos(\phi^i_t), \sin(\phi^i_t), 0 \right]}$.
This step ensures that the UAVs are kept at the desired radial distance from the target.

The swarm control in \eqref{eq:phase-control} runs individually for each UAV, requires only phase information of its two adjacent drones, and is unaffected by the addition or loss of agents at runtime, as proven in \cite{swarm_embedding}. Thus, our technique does not increase computational complexity as the swarm size grows and will continue to work even if UAVs are lost or reassigned to another mission. 

This controller was proposed to encircle a fixed point and has been adapted, in this work, to encircle a moving target. This aspect challenges the system because the UAVs' positions must anticipate the target's motion to avoid keeping the swarm behind the point of interest.

To mitigate this problem, we propose two methods that aim to improve the aforementioned swarm encirclement control law. First, to ensure that the linear speeds of the UAVs in the swarm are greater than the target's speed $\Vector{v}_\text{target}(t)$, the desired nominal angular speed is calculated as $\omega_{z,d} = (k_\omega||\Vector{v}_\text{target}(t)||)/r^i(t)$, where $k_\omega >1$. This way, UAVs can move along while rotating around it.

Next, the encirclement radius is defined dynamically during operation using the target's velocity as follows:
\begin{equation}
    r^i(t) = r_{min} + \max(0,k_r d_r \cos(\alpha)),
    \label{eq:desired-radius}
\end{equation}
where $k_r>0$ is the control gain, $r_{min}$ is the minimum radius of the encirclement, $d_r=||\Vector{v}_\text{target}(t)||\Delta t$,  $\Delta t$ is the time step, and
\begin{equation*}
    \cos(\alpha) = \frac{\left(\Vector{x}^i\left(\phi^i(t+1)\right)\cdot \Vector{v}_\text{target}(t)\right)}{||\Vector{x}^i\left(\phi^i(t+1) \right)|| \text{ }||\Vector{v}_\text{target}(t)||},
\end{equation*}
which corresponds to the cosine of the angle between the UAV's direction and the target's movement, meaning that the radius is maximum when the UAV is ahead of the target and decreases otherwise.
This radial control law prevents the encirclement from wasting time and energy by traveling long distances in the opposite direction to the target's motion. Additionally, the encirclement stretches in the target's direction to compensate for possible delays in the filter estimate, thereby contributing to keeping the target within the encirclement.

\subsection{Swarm State Estimation} \label{sub-sec:swarm-state-estimation}

The first challenge is to estimate the phase separation between agents using only relative measurements, no communication between agents, and no unified global coordinate system. To solve this challenge, a discrete-time Kalman filter is proposed to estimate these strictly positive phase distances.

Let the state vector be defined as $\Vector{x}(t) = \Transpose{[d_{\text{ahead}}, d_{\text{behind}}]} \in [0, 2\pi)^2$, representing the angular separation between the leader and the follower, respectively. 

\begin{figure}
    \centering
    \includegraphics[trim={15cm 5cm 0cm 0},clip,width=\columnwidth]{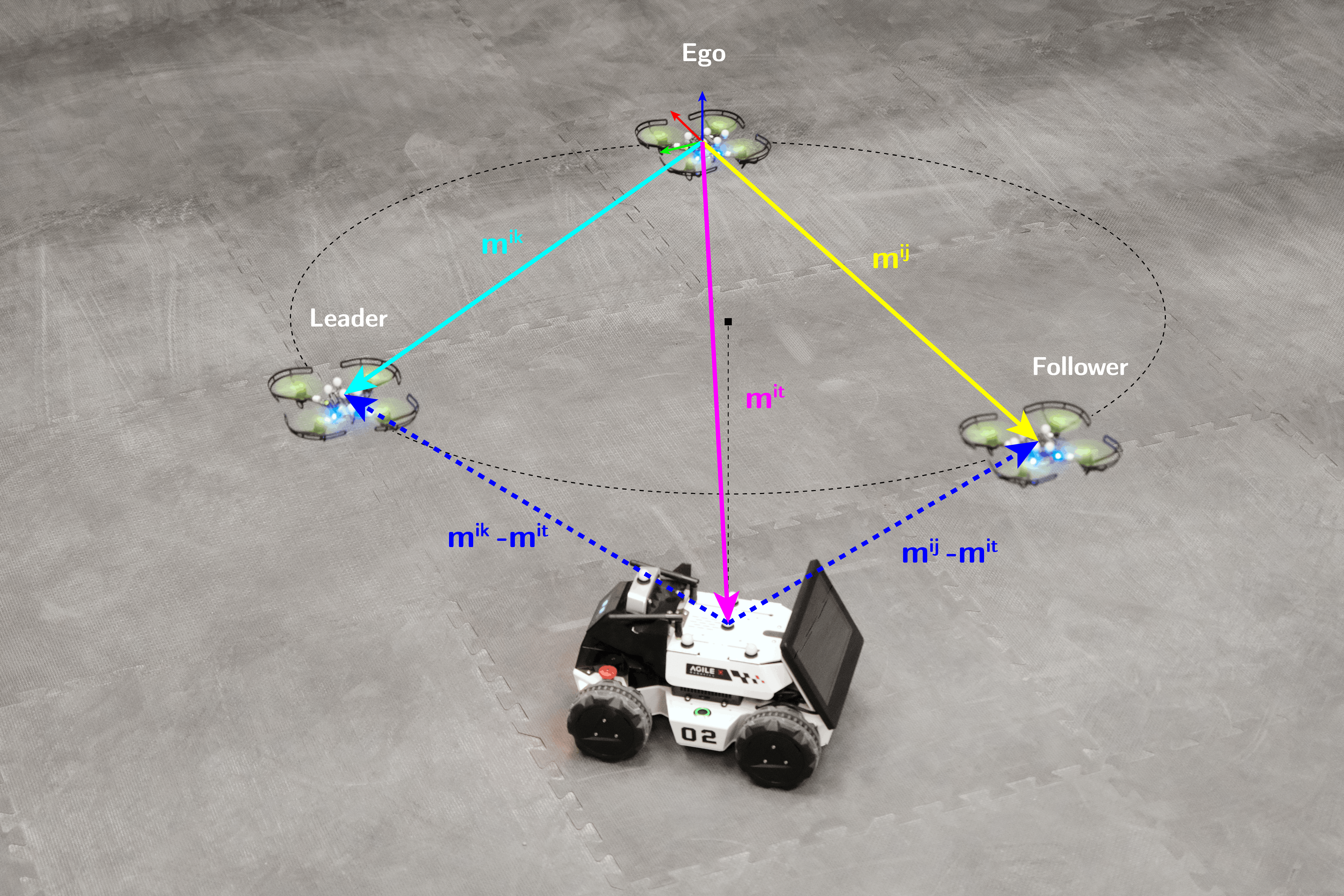}
    \caption{Relative measurements used to estimate the phase separation between the UAVs.}
    \label{fig:swarm-state-estimation}
\end{figure}

\subsubsection{Prediction Model}
Assuming the nominal angular velocities of all UAVs are identical during stable encirclement, the inter-agent phase differences are expected to be constant. Thus, the system kinematics is modeled as a random walk process:
\begin{align}
    \Vector{x}(t+1) &= \Vector{x}(t) \\
    P(t+1) &= P(t) + Q \Delta t
\end{align}
where $P \in \Real^{2\times2}$ is the state covariance matrix, and $Q = \text{diag}(q_{\text{ahead}}, q_{\text{behind}})$ is the process noise covariance matrix. The process noise models the formation flexibility, accounting for minor speed fluctuations among the independent agents.

\subsubsection{Measurement Model}
As illustrated in \figurename~\ref{fig:swarm-state-estimation}, the ego agent obtains 3D relative position vectors in its local body frame: $\Vector{m}^{it}$ (to the central target), $\Vector{m}^{ik}$ (to the leader), and $\Vector{m}^{ij}$ (to the follower). 


To extract the relative angles, these measurements are first transformed into a globally aligned coordinate frame using the ego drone's current rotation matrix, $R^i \in \text{SO}(3)$, which comes from the odometry system. Then, the radial vectors originating from the target to each respective drone are constructed:
\begin{align}
    \mathbf{v}^i &= -R^i \Vector{m}^{it}, \\
    \mathbf{v}^k &= R^i \Vector{m}^{ik} - R^i\Vector{m}^{it}, \\
    \mathbf{v}^j &= R^i \Vector{m}^{ij} - R^i\Vector{m}^{it}.
\end{align}

The measurement vector $\Vector{z}(t) \in \Real^2$ consists of the counter-clockwise angles between these radial vectors projected onto the 2D plane:
\begin{equation}
    \mathbf{z}(t) = \left[
    \begin{smallmatrix}
        \angle(\Vector{v}^i, \Vector{v}^k) \pmod{2\pi} \\
        \angle(\Vector{v}^j, \Vector{v}^i) \pmod{2\pi}
    \end{smallmatrix}\right].
\end{equation}
The angles are computed using $\text{arctan2}$ on the 2D cross and dot products of the vectors.

The measurement mapping is the identity matrix $H = I^{2\times2}$. Because the state represents angles on a circular domain, the equations must be adapted to handle $2\pi$ wraparound. To guarantee that the filter computes the shortest path error, the innovation sequence $\Vector{y}(t)$ is wrapped to the interval $[-\pi, \pi)$.

\subsection{Target State Estimation} \label{sub-sec:target-state-estimation}

The next challenge is to find the target in the arbitrarily defined UAV coordinate system. \textbf{Target 1} is modeled as a unicycle operating on an estimated ground plane altitude $z_g$. To estimate the unknown trajectory, the geometric state is augmented with the unicycle's control parameters and the ground height, as there is no communication between the target and the UAV. The extended state vector is defined on the manifold $\mathcal{M} = \text{SE}(2) \times \Real^3$:
\begin{align}
    \Vector{x}(t) &= \left(\mathcal{X}(t), \omega(t), v(t), z_g(t)\right) \\
    \mathcal{X}(t) &= \left[
    \begin{smallmatrix}
        R(\theta(t)) & \Vector{p}(t) \\
        \Vector{0}_{1\times 2} & 1
    \end{smallmatrix}\right],
\end{align}
where $R(\theta^t(t)) \in \text{SO}(2)$ is the planar rotation matrix, $\Vector{p}(t) = \Transpose{[x, y]} \in \Real^2$ is the 2D position, $(\omega(t) \in \Real$, $v(t)) \in \Real$, and $z_g(t) \in \Real$ represents the angular and linear velocities, and the ground elevation of the target.

Adopting a constant velocity model for the prediction step ($\dot{\omega}^v \approx 0, \dot{v}^v \approx 0, \dot{z}^g \approx 0$), the continuous-time dynamics are given by:
\begin{equation*}
    \dot{R} = R (\omega)^\wedge,
    \dot{\Vector{p}} = R \Vector{v},
    \dot{\omega} = w^\omega,
    \dot{v} = w^v, 
    \dot{z}_g = w^z,
\end{equation*}
where $\Vector{v} = \Transpose{[v(t), 0]}$, $(\cdot)^\wedge$ denotes the skew-symmetric operator for the Lie algebra $\mathfrak{so}(2)$, and $w^\omega, w^v, w^z$ represent Gaussian process noise accounting for unmodeled accelerations and terrain variations.

\subsubsection{Virtual Measurement Formulation}
The tracking UAV obtains a relative 3D position vector of the target, denoted as $\Vector{y}_r \in \Real^3$, expressed in the UAV's local body frame. A \textit{virtual global measurement} $\Vector{y} \in \Real^3$ is constructed by transforming the relative observation into the fixed frame defined for the UAV:
\begin{equation}
    \Vector{y}(t) = R^i(t) \Vector{y}_r(t) + \Vector{p}^i(t) \approx \left[
    \begin{smallmatrix}
        \mathbf{p}(t) \\
        z_g(t)
    \end{smallmatrix} \right] + \mathbf{n}(t),
\label{eq:virtual_meas}
\end{equation}
where $R^i(t) \in \text{SO}(3)$ and $\Vector{p}^i(t) \in \Real^3$ are the UAV's orientation and position. The measurement noise $\Vector{n}(t) = R^i(t) \Vector{n}_r$ represents the raw sensor covariance $N_r$ rotated into the fixed frame, yielding an effective measurement covariance $N(t) = R^i(t) N_r \Transpose{R^i(t)}$.

\subsubsection{Invariant Extended Kalman Filter}
A Left-Invariant Extended Kalman Filter (LIEKF) on $\text{SE}(2)$ is employed to estimate the target's state. The left-invariant error $\eta(t) \in \text{SE}(2)$ encapsulates the planar discrepancy in the target's local frame. For compactness, the time dependence is omitted in the following equations. Using the Lie algebra logarithm map, the full $6$-dimensional error state vector $\boldsymbol{\xi}^t \in \Real^6$ is defined as:
\begin{equation}
    \label{eq:error-state-vector}
    \boldsymbol{\xi} =
    \Transpose{[\xi_x, \xi_y, \xi_\theta, \xi_\omega, \xi_v, \xi_z]} = \left[
    \begin{smallmatrix}
        \log(\eta)^\vee \\
        \hat{\omega} - \omega \\
        \hat{v} - v \\
        \hat{z}_g - z_g
    \end{smallmatrix}\right],
\end{equation}
where $(\cdot)^\vee$ is the inverse operation of $(\cdot)^\wedge$.
The linearized error dynamics $\dot{\boldsymbol{\xi}} = A \boldsymbol{\xi} + \Vector{w}$ yield the following continuous-time Jacobian $A$:
\begin{equation}
    A = \left[
    \begin{smallmatrix} 
    0 & \hat{\omega} & 0 & 0 & 1 & 0 \\
    -\hat{\omega} & 0 & \hat{v} & 0 & 0 & 0 \\
    0 & 0 & 0 & 1 & 0 & 0 \\
    0 & 0 & 0 & 0 & 0 & 0 \\
    0 & 0 & 0 & 0 & 0 & 0 \\
    0 & 0 & 0 & 0 & 0 & 0 
    \end{smallmatrix}
    \right].
\end{equation}

Leveraging the virtual measurement $\Vector{y}$, the innovation vector $\Vector{z} \in \Real^3$ isolates the planar (body frame) and vertical (global frame) residuals:
\begin{equation}
    \Vector{z} = \left[
    \begin{smallmatrix}
        \Transpose{\hat{R}} \left( \Vector{y}_{xy} - \hat{\Vector{p}} \right) \\
        y_z - \hat{z}_g
    \end{smallmatrix} \right],
\end{equation}
yielding a constant measurement Jacobian $H \in \Real^{3 \times 6}$:
\begin{equation}
    H = 
    \left[
    \begin{smallmatrix} 
    1 & 0 & 0 & 0 & 0 & 0 \\ 
    0 & 1 & 0 & 0 & 0 & 0 \\
    0 & 0 & 0 & 0 & 0 & 1 
    \end{smallmatrix} \right].
\end{equation}
Following the standard computation of the Kalman gain $K$, the additive parameters are updated linearly ($\hat{v} \leftarrow \hat{v}_t + \xi^v$, etc.), while the planar pose undergoes a left-invariant geometric update $\hat{\mathcal{X}} \leftarrow \hat{\mathcal{X}} \exp\left(\boldsymbol{\xi}^{\wedge}_{\mathfrak{se}(2)}\right)$.

Based on the rank condition of the observability matrix, the unicycle system is locally weakly observable provided $v_t \neq 0$. When the target stops, the heading $\theta$ becomes mathematically unobservable, as rotation generates no positional displacement. To prevent heading covariance explosion during stationary periods, a Zero Velocity Update (ZUPT) architecture is implemented. If the estimated speed $|\hat{v}_t|$ falls below a threshold $\epsilon$, the velocity is clamped to zero and the continuous position noise variance is scaled down, while the heading process noise remains active to capture potential in-place rotations. 

While the unicycle-based LIEKF is mainly designed for non-holonomic ground vehicles, the same filter architecture is also deployed to estimate the state of \textbf{Target 2}. Unlike a ground rover, a UAV is a holonomic platform capable of independent lateral translations (side-slip) and rapid vertical accelerations, which inherently violate the strict non-holonomic constraints of the unicycle model (i.e., $\dot{\Vector{p}} = R \Vector{v}$). To compensate for this kinematic mismatch without increasing the system state complexity, this work adopts a covariance relaxation strategy. Specifically, the continuous-time process noise components corresponding to the planar positions ($x, y$) and the altitude ($z_g$) in the $Q$ matrix are substantially inflated. By increasing the modeled uncertainty of the positional states, the filter is explicitly instructed to trust the underlying prediction model less and rely more heavily on the incoming virtual measurements. This allows the mathematically simpler unicycle filter to fluidly absorb unmodeled aerial dynamics, maintaining a reasonable tracking performance while keeping the computational architecture unified across heterogeneous targets.

\section{EXPERIMENTS} \label{sec:experiments}
\subsection{Experimental Setup}
The proposed pipeline was validated through real experiments. We used four Crazyflies quadcopters and one Limo as \textbf{Target 1}, as illustrated in \figurename~\ref{fig:scenario}.
Due to space constraints in the testing area, three quadcopters were used as \textbf{Defenders}, and the fourth one was used as \textbf{Target 2}.
The quadcopters used a low-level position controller that received only waypoints. Therefore, the quadcopters did not perform aggressive maneuvers with dynamic attitude changes. They always moved parallel to the ground, resembling a unicycle robot moving on different horizontal planes.
However, the pipeline can be extended to accommodate more agents, as mentioned in Section~\ref{sub-sec:swarm-control}. Each quadcopter had its own ROS2 node that received noisy position measurements and calculated the estimated target position and angular separation to its neighbors, along with its own control commands. 

Experiments were conducted in a test area equipped with a Vicon tracking system.
Because Crazyflies lack the payload capacity to attach sensors, relative distance measurements were obtained from the Vicon system. To emulate the behavior of a ranging sensor, such as LiDARs, each relative distance was transformed into its respective drone's initial frame, and zero-mean Gaussian noise with a variance of $0.02$~m was added. We assumed that the UAVs had an unlimited field of view and did not suffer from occlusion.

\subsection{Experiments}


The filters predicted the states at $100$~Hz, which were corrected at $10$~Hz. The ZUPT threshold was set to $0.005$~m/s. The swarm control gains were $r_{min}=0.5$~m, $k_\phi=2.1$, $k_r=15.0$, $k_\omega=1.1$. A human teleoperated the \textbf{Target 1} with maximum linear and angular speeds of $0.36$~m/s and $1.60$~rad/s, respectively. 
We tested the entire pipeline with all components working together. In Phase 1, the three \textbf{Defenders} are initialized by encircling \textbf{Target 1}. The \textit{Yellow Zone} was set at a radius of $3.5$~m. Later, \textbf{Target 2} started the attack by flying towards \textbf{Target 1}. Once in a while, \textbf{Target 1}, random values uniformly distributed from $-0.3$~m to $0.3$~m were injected into \textbf{Target 2}'s $z$ coordinate to simulate evasive maneuvers and evaluate the estimation of this coordinate. The \textbf{Defenders} detected the attacker and navigated to its position using the flocking algorithm proposed in~\cite{Olfati-Saber_flocking_2006}. The \textbf{Defenders} then began encircling \textbf{Target 2}. The \textit{Red Zone} radius was $r_{\text{safe}} = 1$~m. Upon the attacker entering this zone, the drones initiated the neutralization protocol and reduced the encirclement radius to collide with the attacker. The results presented in the next section are separated according to the same pipeline order.


\section{RESULTS AND DISCUSSION} \label{sec:results}

\subsection{Phase 1}
\begin{figure}
    \centering
    \includegraphics[trim={0.45cm 0.45cm 0.45cm 0.35cm}, clip, width=\columnwidth]{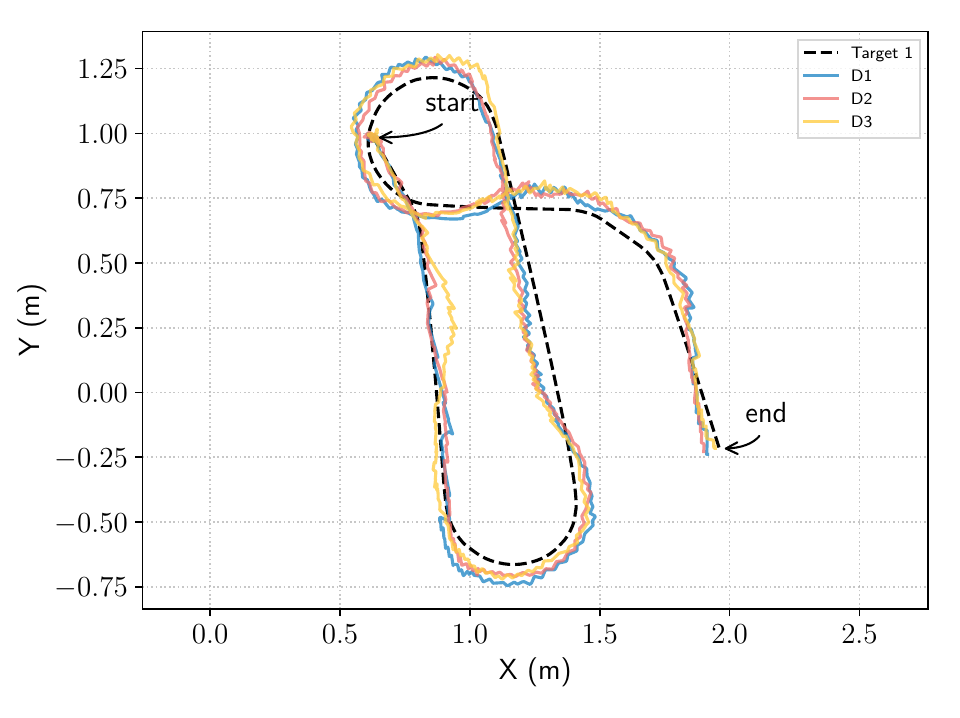}
    \caption{Ground truth (black dashed line) and \textbf{Defenders}' estimated trajectories while defending the ground vehicle target.}
    \label{fig:trajectory-following-limo}
\end{figure}

In this section, we present the results obtained during Phase 1 of the pipeline. 
\figurename~\ref{fig:trajectory-following-limo} shows the ground truth trajectory of \textbf{Target 1} and the trajectories  estimated by each defender. The root-mean-square-error (RMSE) of the 3D position for D1, D2 and D3 were $0.035$~m, $0.032$~m and $0.035$~m. These are remarkable results given the system challenge, i.e., estimating all states and control inputs in a highly dynamic environment. 

\begin{figure}
    \centering
    \includegraphics[trim={0.3cm 0.45cm 0.35cm 0.35cm}, clip, width=\columnwidth]{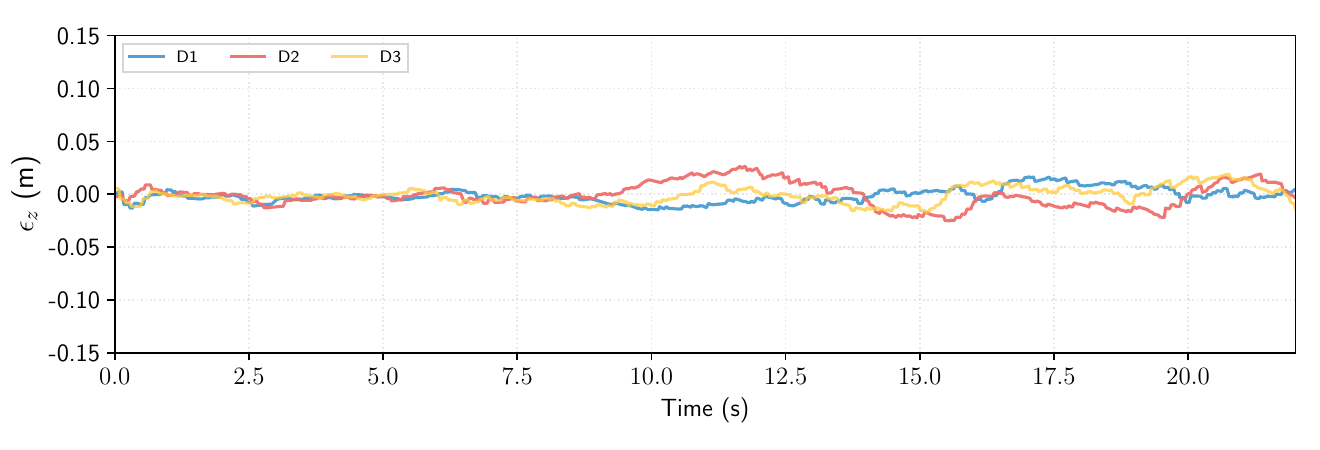}
    \caption{Error in time of the $z$ coordinate estimated by the \textbf{Defenders} during Phase 1.}
    \label{fig:error-z-limo}
\end{figure}

The error $\epsilon_z$ in the $z$ coordinate estimated by \textbf{Defenders} as a function of time is presented in \figurename~\ref{fig:error-z-limo}. The error is bounded by $0.03$~m, demonstrating the consistency and precision of the estimation throughout this part of the experiment.

\begin{figure}
    \centering
    \includegraphics[trim={0.3cm 0cm 0.45cm 0.35cm}, clip, width=\columnwidth]{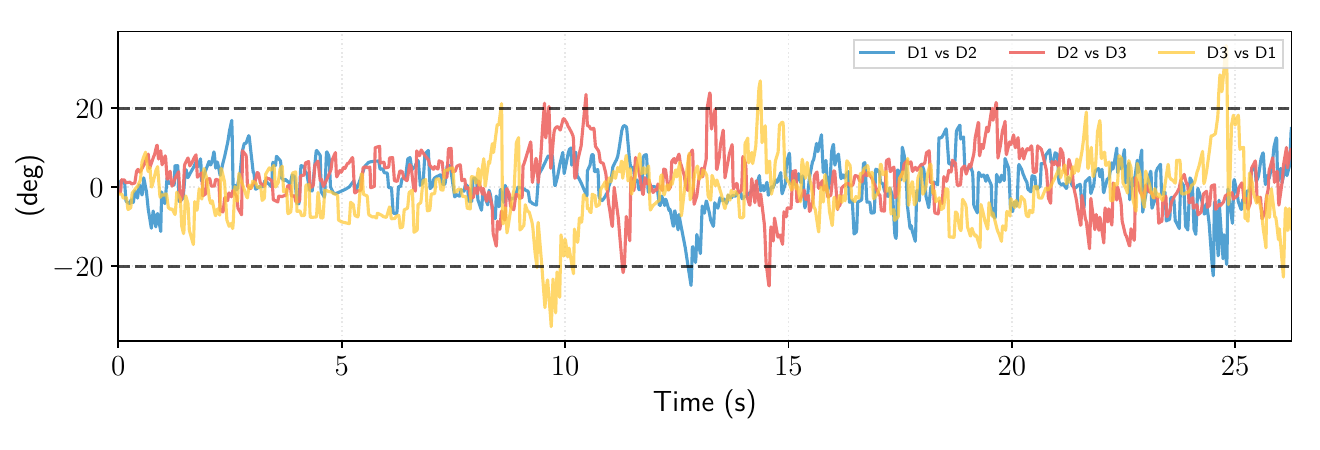}
    \caption{Estimated phase difference error while defending \textbf{Target 1}.}
    \label{fig:phase-difference-filter-limo}
\end{figure}

While encircling \textbf{Target 1}, the swarm was exposed to random dynamics, including variable speeds. In some maneuvers, the phase difference estimation error exceeded $20^\circ$ as shown in \figurename~\ref{fig:phase-difference-filter-limo}. These results illustrate how vehicle dynamics affected the estimation performed with only relative measurements. The RMSE values were $5.94^\circ$ (D1-D2), $7.02^\circ$ (D2-D3) and $7.90^\circ$ (D3-D1).

\begin{figure}
    \centering
    \includegraphics[trim={0.25cm 0cm 0.45cm 0.35cm}, clip, width=\columnwidth]{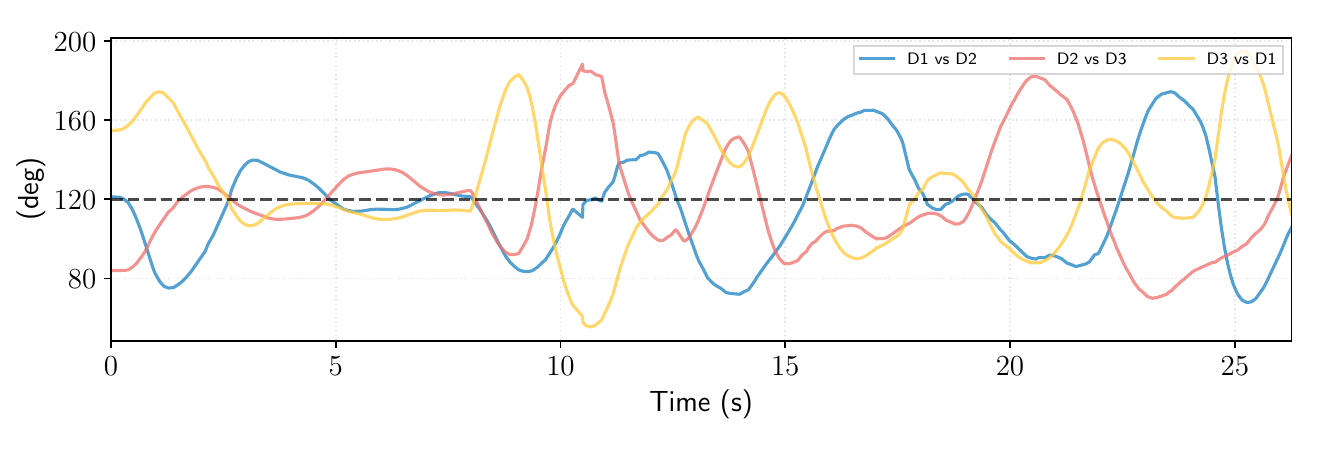}
    \caption{True phase difference while defending \textbf{Target 1}.}
    \label{fig:phase-difference-controller-limo}
\end{figure}

\begin{figure}
    \centering
    \includegraphics[width=\columnwidth]{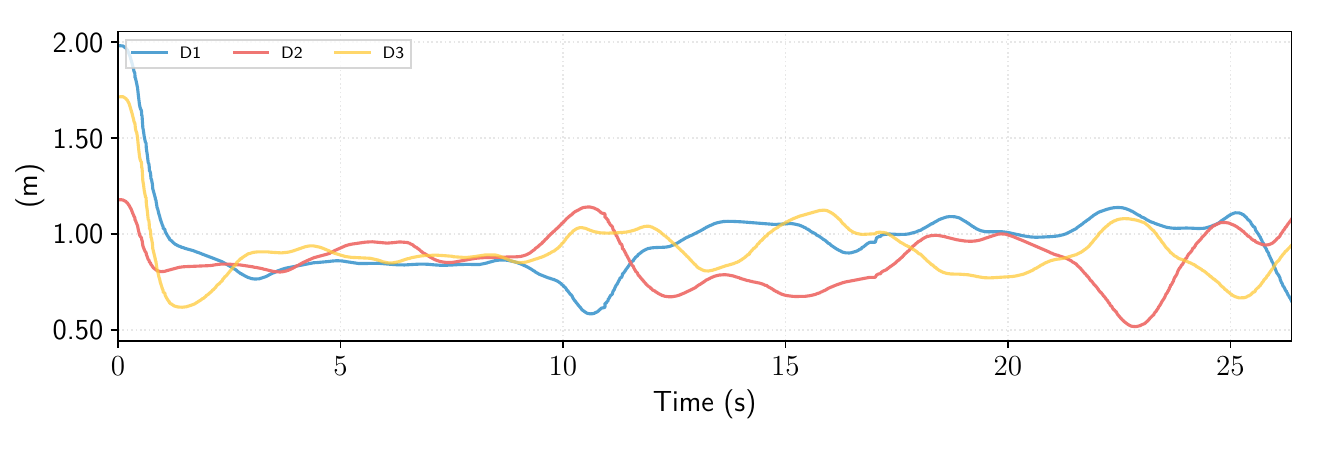}
    \caption{Distance between each \textbf{Defender} and its leader while encircling \textbf{Target 1}}
    \label{fig:dist-leader-following-limo}
\end{figure}

Also, in \figurename~\ref{fig:phase-difference-controller-limo}, we plotted the actual phase differences. The minimum and maximum values were $60^\circ$ and $180^\circ$. Even with this variation, the average values were $121.79^\circ$ (D1-D2), $126.52^\circ$ (D2-D3), and $145.51^\circ$ (D3-D1). The overall results were close to the desired separation, i.e., $120^\circ$. Despite noise in the estimation, the controller successfully maintained the UAVs free of collisions, even at the minimum and maximum angular separation. This is demonstrated by \figurename~\ref{fig:dist-leader-following-limo}, in which the minimum inter-UAV distance was $0.5$~m (minimum collision distance between Crazyflies is $0.17$~m).

\begin{figure}
    \centering
    \includegraphics[trim={0.25cm 0cm 0.45cm 0.35cm}, clip, width=\columnwidth]{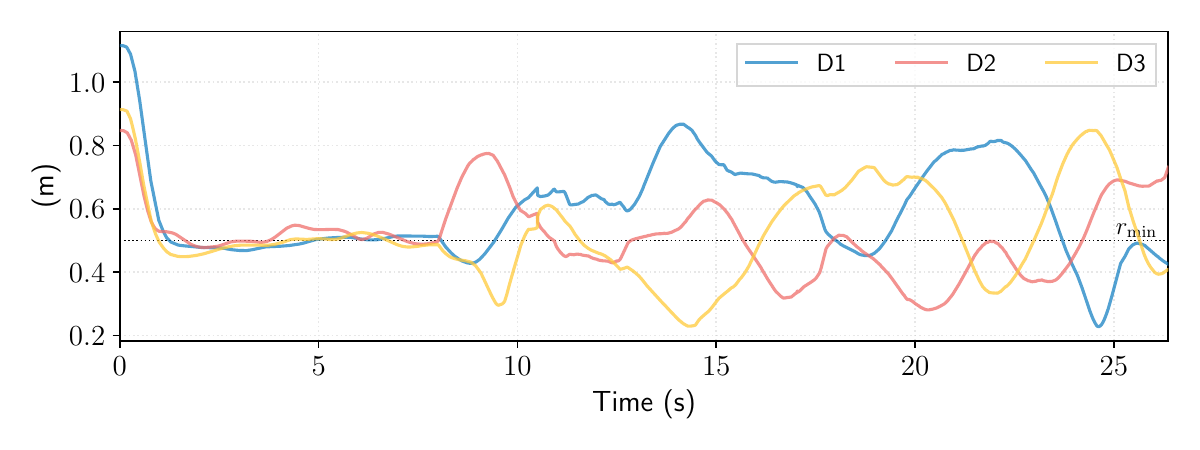}
    \caption{Encirclement radii of the \textbf{Defenders} while encircling \textbf{Target 1}.}
    \label{fig:radii-target-1}
\end{figure}

The adaptive encirclement radii while encircling \textbf{Target 1} are shown in \figurename~\ref{fig:radii-target-1}. Due to the target's irregular velocity profile, the radius varied from $0.22$~m to $0.85$~m. Although the desired radius in \eqref{eq:desired-radius} is greater than or equal to $r_{\text{min}}$, and the desired drone position was calculated using \eqref{eq:radial-position}, the actual radius occasionally dropped below this bound. This occurred because the proposed pipeline does not include an explicit radial correction to enforce the equality between the desired and actual radii.

\begin{figure}
    \centering
    \includegraphics[trim={0.25cm 0cm 0.45cm 0.35cm}, clip, width=\columnwidth]{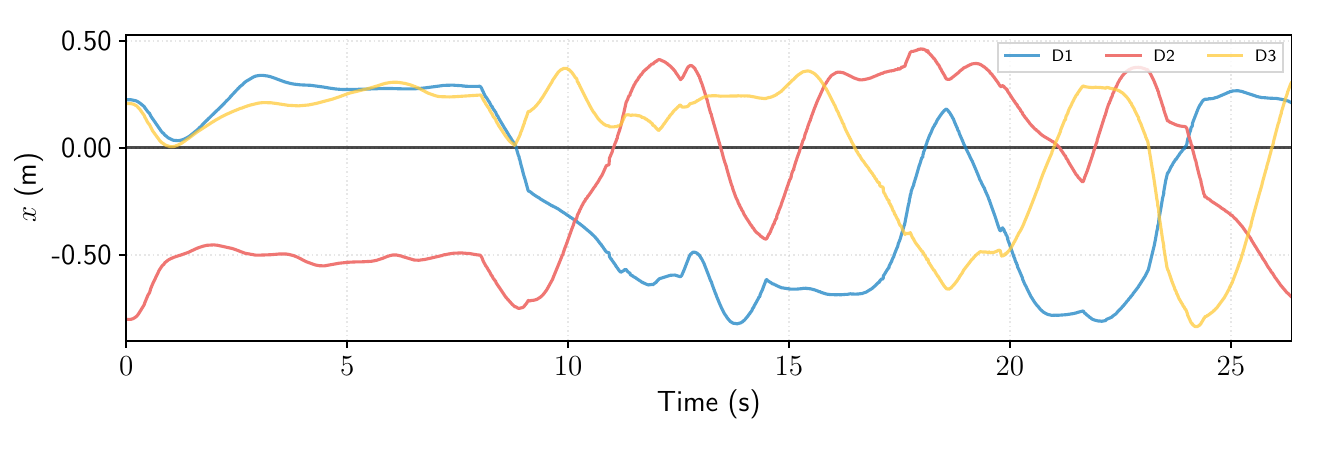}
    \caption{$x$ component of \textbf{Defenders}' positions in time while encircling \textbf{Target 1}.}
    \label{fig:x-component-1}
\end{figure}

\figurename~\ref{fig:x-component-1} shows the \textbf{Defenders}' $x$-coordinate expressed in the \textbf{Target 1}'s body frame. Throughout the experiment, at least one drone maintained a positive $x$-coordinate, i.e., remained ahead of the target,  demonstrating the effectiveness of the adaptive encirclement.

\subsection{Transition}
\begin{figure}
    \centering
    \includegraphics[trim={5cm 1.5cm 4cm 2cm}, clip, width=\columnwidth]{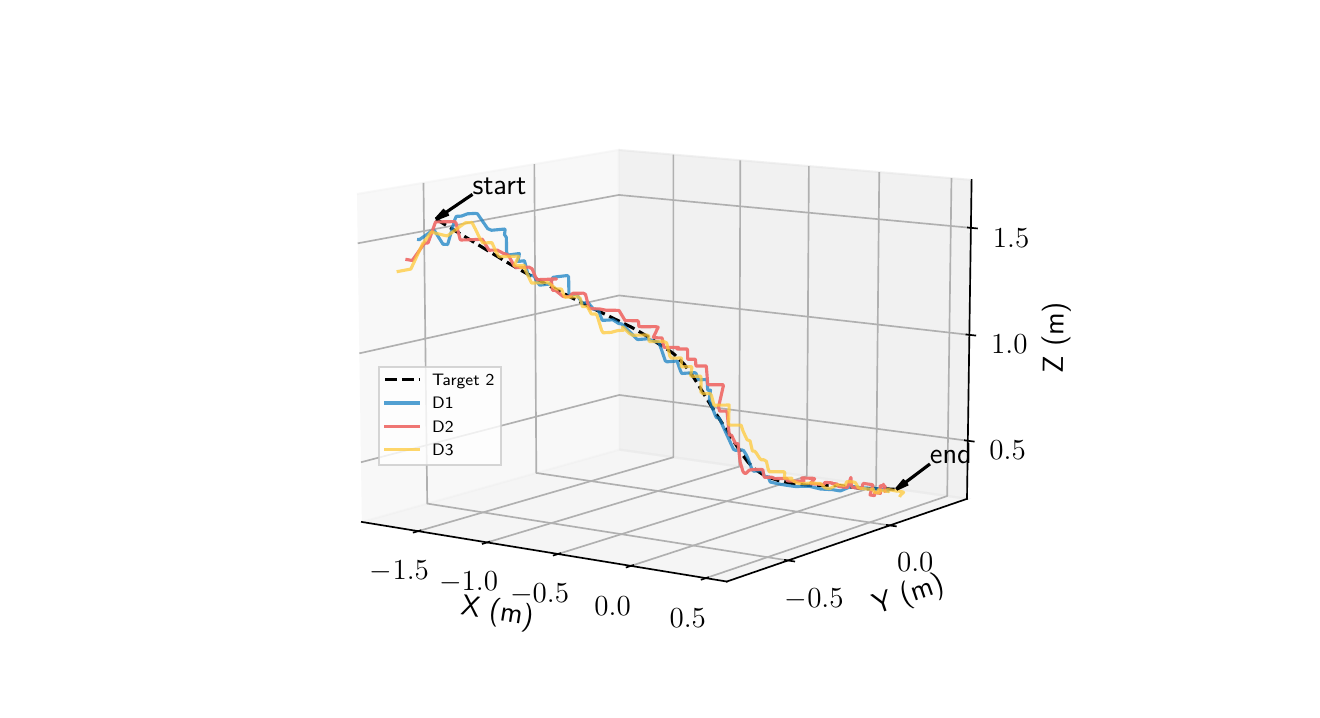}
    \caption{Ground truth (black dashed line) and \textbf{Defenders}' estimated trajectories while transitioning from \textbf{Target 1} to \textbf{Target 2}.}
    \label{fig:trajectories-transition-phase}
\end{figure}

During the Transition phase, the filters were initialized upon detection of \textbf{Target 2} and had little time to warm up. Even so, they kept the estimates close enough to the ground truth to succeed in the transition to Phase 2, as shown in \figurename~\ref{fig:trajectories-transition-phase}. The 3D RMSE for D1, D2, and D3 were $0.107$~m, $0.050$~m and $0.053$~m, respectively.

\subsection{Phase 2}

\begin{figure}
    \centering
    \includegraphics[trim={5cm 1.5cm 4cm 2cm}, clip, width=\columnwidth]{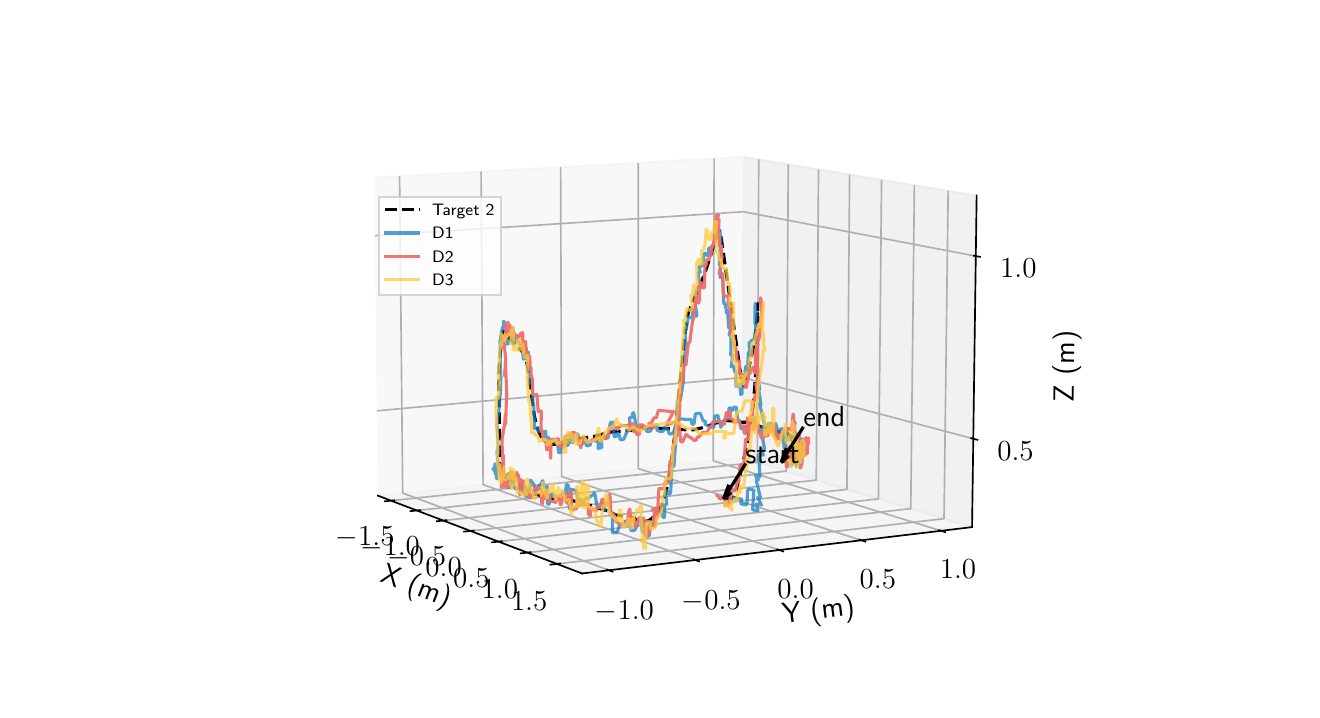}
    \caption{Ground truth (black dashed line) and \textbf{Defenders}' estimated trajectories while encircling \textbf{Target 2}.}
    \label{fig:trajectory-following-drone}
\end{figure}

\begin{figure}
    \centering
    \includegraphics[trim={0.25cm 0.45cm 0.35cm 0.35cm}, clip, width=\columnwidth]{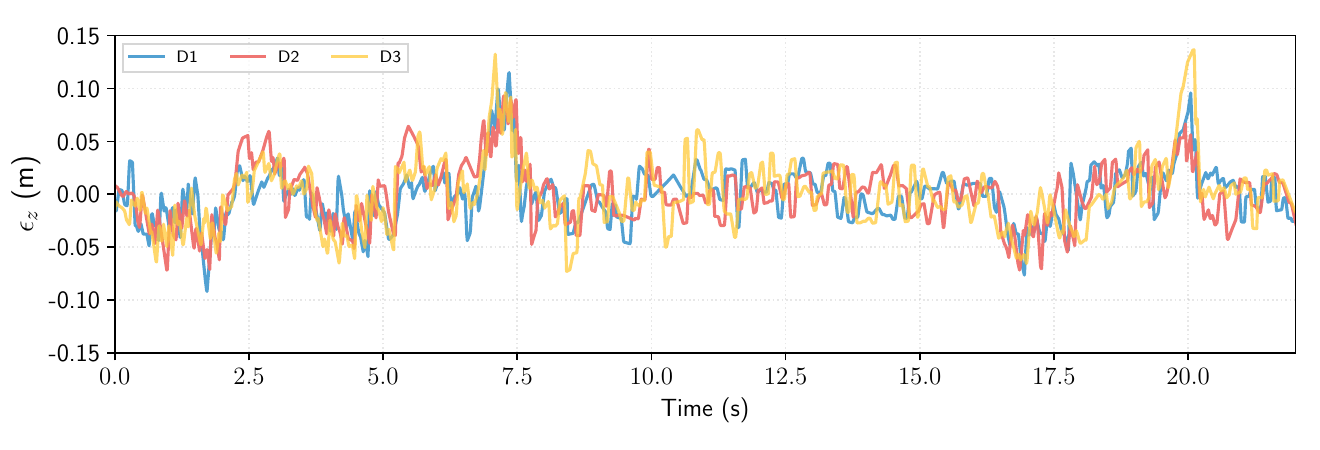}
    \caption{Error in $z$ coordinate estimated by the \textbf{Defenders} during Phase 1.}
    \label{fig:error-z-drone}
\end{figure}

The ground truth 3D trajectory executed by \textbf{Target 2} and each \textbf{Defender}'s estimation are illustrated in \figurename~\ref{fig:trajectory-following-drone}. The three \textbf{Defenders}' estimations remained consistent throughout the trajectory and close to the ground truth, with 3D RMSE of $0.029$~m, $0.028$~m, and $0.037$~m for D1, D2, and D3. 
\figurename~\ref{fig:error-z-drone} shows the error $\epsilon_z$ in the $z$ coordinate estimated by the \textbf{Defenders}, which was noisier than that of \textbf{Target 1} (see ~\figurename~\ref{fig:error-z-limo}) because our model does not account for velocity measurements in the $z$-axis, making predictions less accurate for targets moving along this axis. Even under that assumption, the filter maintained the error bounded within $\pm0.10$~m most of the time and successfully tracked \textbf{Target 2}.

\begin{figure}
    \centering
    \includegraphics[trim={0.3cm 0cm 0.45cm 0.35cm}, clip, width=\columnwidth]{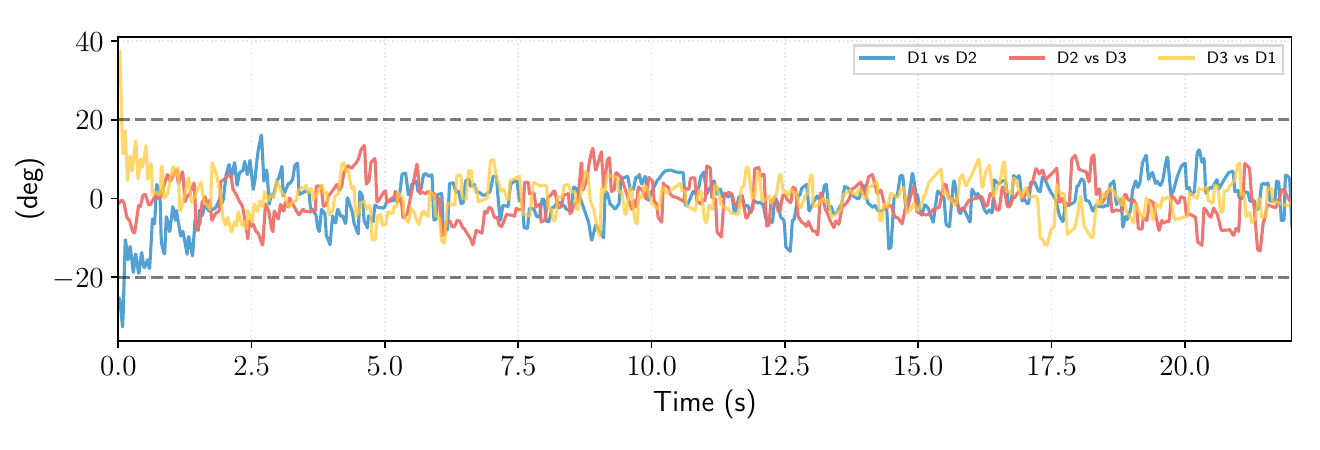}
    \caption{Estimated phase difference error while encircling \textbf{Target 2}.}
    \label{fig:phase-difference-filter-drone}
\end{figure}

Similarly, we plotted the phase difference estimation error in \figurename~\ref{fig:phase-difference-filter-drone}. The error was less than $20^\circ$ for most of the trajectory. The RMSE values compared against the ground truth were $5.56^\circ$ (D1-D2), $4.80^\circ$ (D2-D3), and $4.63^\circ$ (D3-D1). textbf{Target 2 }'s exhibited variation and randomness, but the filters successfully estimated the phase differences with tighter variances.

\begin{figure}
    \centering
    \includegraphics[trim={0.25cm 0cm 0.45cm 0.35cm}, clip, width=\columnwidth]{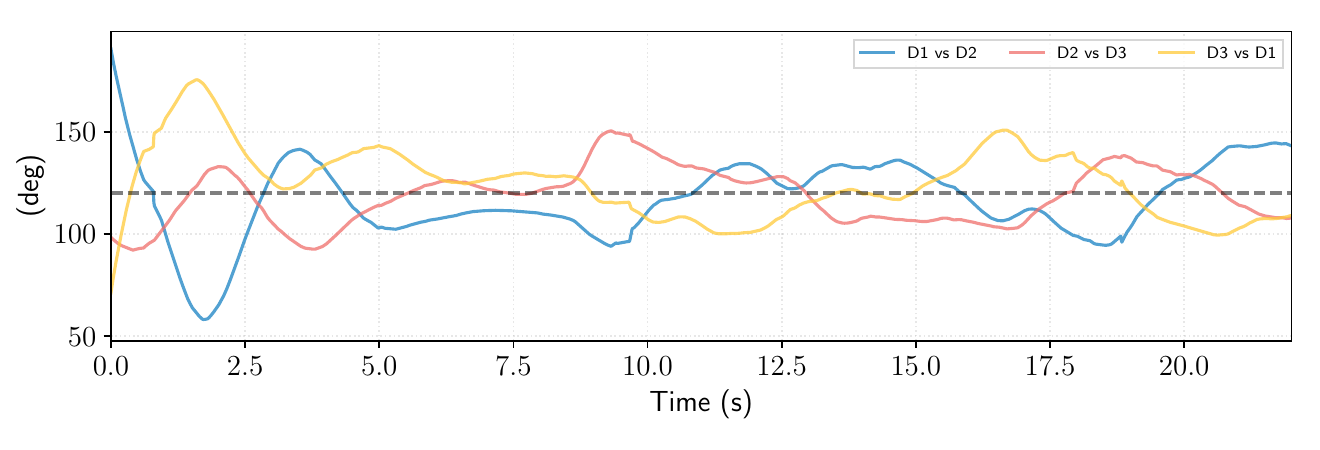}
    \caption{True phase difference while while encircling \textbf{Target 2}.}
    \label{fig:phase-difference-controller-drone}
\end{figure}

 The actual phase differences extracted from the tracking system are presented in \figurename~\ref{fig:phase-difference-controller-drone}. Its average values were  $120.19^\circ$ (D1-D2), $121.02^\circ$ (D2-D3), and $133.93^\circ$ (D3-D1). 
 The phase controller performed better than in Phase 1 due to better estimation of the phase difference. It maintained the angular separation about the desired value, meaning that collisions were avoided.

\begin{figure}
    \centering
    \includegraphics[trim={0.25cm 0cm 0.45cm 0.35cm}, clip, width=\columnwidth]{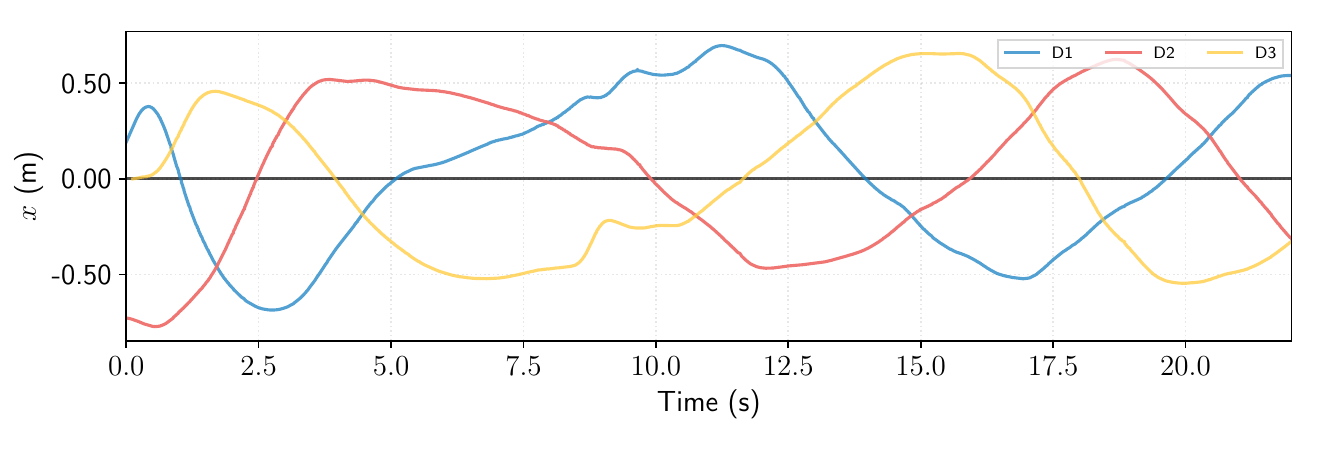}
    \caption{$x$ component of \textbf{Defenders}' positions in time while encircling \textbf{Target 2}.}
    \label{fig:x-component-2}
\end{figure}

We also present the effectiveness of the proposed adaptive encirclement radii when the \textbf{Defenders} encircled \textbf{Target 2}. \figurename~\ref{fig:x-component-2} also shows that at least one drone was successfully kept in front of \textbf{Target 2}.

\subsection{Neutralization}

\begin{figure}
    \centering
    \includegraphics[trim={0.5cm 0.15cm 0.5cm 0.25cm},clip,width=\columnwidth]{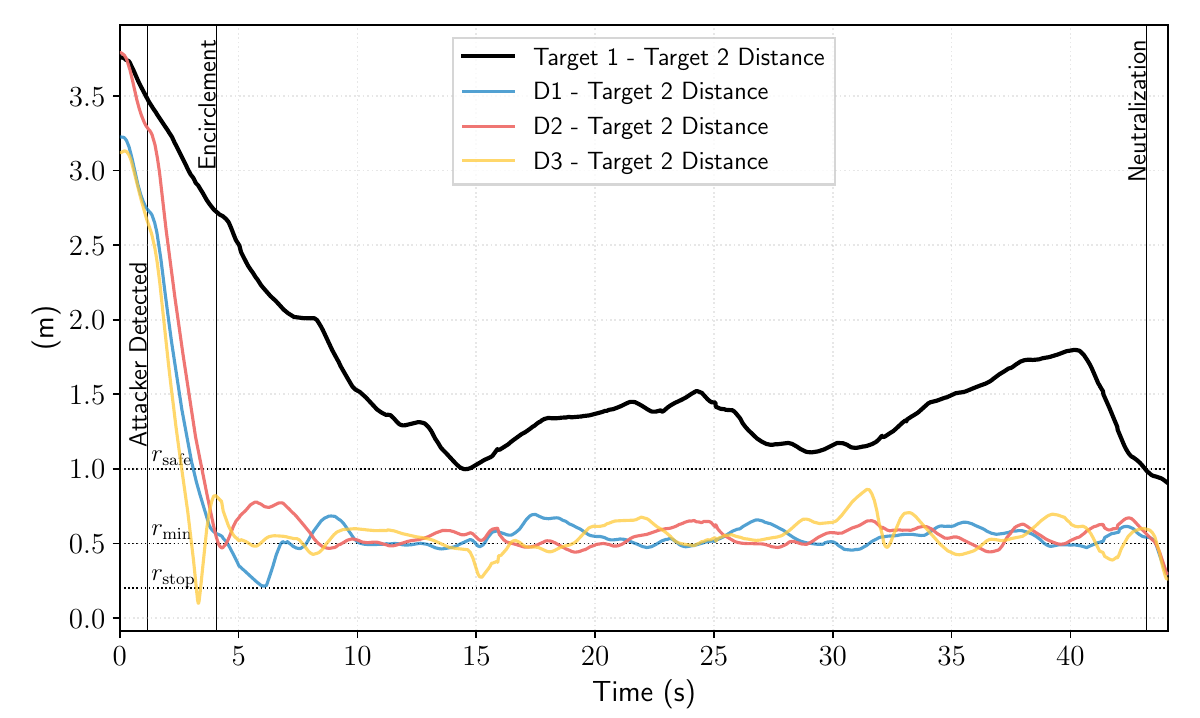}
    \caption{Distances from \textbf{Defenders} and \textbf{Target 1} to \textbf{Target 2}, from the beginning of the attack until neutralization.}
    \label{fig:transition-phase2-neutralization}
\end{figure}

\figurename~\ref{fig:transition-phase2-neutralization} shows the distance from \textbf{Target 1} and \textbf{Defenders} to \textbf{Target 2}, from the beginning of the attack to neutralization, \textit{i.e.,} excluding Phase 1. After the unauthorized drone was detected, the \textbf{Defenders} quickly switched to Transition Phase, and the distance between them and \textbf{Target 2} decreased, as did the distance between the two targets. Phase 2 began at $4.1$~s. \textbf{Target 1} continued to move while the \textbf{Defenders} maintained their position relative to \textbf{Target 2} at around $r_{\text{min}}$
Once \textbf{Target 2} trespassed the \textit{Red Zone} defined by $r_\mathrm{safe}$, the defenders activated the Neutralization Phase ($43.6~s$), where the radius of the encirclement collapsed, generating a collision among the UAVs. To avoid damaging the robots, we defined a stop radius ($r_{\text{stop}}$) of $0.02$~m, and the drones landed upon reaching this value.

\subsection{Discussion}

In the experiments, we observed that state estimation delays significantly affect encirclement and swarm coordination. To mitigate this, the radial gain $k_r$ was tuned to prevent the swarm from lagging behind the target during acceleration.


The experiment demonstrated successful end-to-end execution. Phase transitions occurred as expected. The phase separation controller stabilized the swarm around the tracked target, and
the adaptive encirclement maintained the targets near the formation center.


The Transition Phase used a flocking algorithm; however, any collision-free swarm navigation technique 
could have been employed. 
The angular separation filter assumes identical angular velocities for all drones in the swarm. Yet these velocities depend on locally estimated target velocity, which may differ across \textbf{Defenders} in a decentralized setup. This limitation was evident when tracking \textbf{Target 1}, whose varying velocity led to larger phase estimation errors compared to \textbf{Target 2} and lower phase controller performance.


The proposed filters can be adapted to different kinematic models (e.g., fast and dynamic maneuvers), sensor modalities, and noise characteristics. Filter and controller parameters (e.g., $k_r$) can also be tuned (online or offline) to match mission conditions, such as slower target motion. 
Finally, the approach is compatible with any range and bearing sensor that provides relative measurements, making it adaptable to different UAV platforms and sensing technologies, including LiDAR, radar, and vision systems.


\section{CONCLUSION} \label{sec:conclusion}

Our work presented and successfully tested a complete pipeline for encircling and protecting a ground vehicle against UAV attacks using a swarm of UAVs, under denied global position information and communication.
Additionally, to guarantee that the targets were kept within the encirclement, a novel adaptive swarm controller was proposed to adapt the angular velocity and radius of the encirclement using the estimated target's velocity.
For future work, we aim to extend the adaptive controller to handle multi-attacker scenarios, requiring the swarm to dynamically partition itself to protect the ground vehicle from simultaneous threats. Also, we plan to investigate more accurate kinematic models for aerial targets to be used within the IEKF framework.  





\balance

\bibliographystyle{IEEEtran}
\bibliography{refs}

\end{document}